\documentclass[10pt,twocolumn,letterpaper]{article}

\usepackage[pagenumbers]{cvpr} 

\usepackage{comment}

\usepackage{booktabs}
\usepackage{multirow}
\usepackage{graphicx}
\usepackage[normalem]{ulem}
\useunder{\uline}{\ul}{}

\usepackage{algorithm,algpseudocode}

\usepackage{xspace}

\usepackage[dvipsnames]{xcolor}

\usepackage{bm}

\definecolor{cvprblue}{rgb}{0.21,0.49,0.74}
\usepackage[pagebackref,breaklinks,colorlinks,citecolor=cvprblue]{hyperref}

\def\paperID{7387} 
\def\confName{CVPR}
\def\confYear{2024}

\title{One Prompt Word is Enough to Boost Adversarial Robustness for Pre-trained Vision-Language Models}

\author{Lin Li\textsuperscript{1}\thanks{equal contribution}, Haoyan Guan\textsuperscript{1}\footnotemark[1], Jianing Qiu\textsuperscript{2}, Michael Spratling\textsuperscript{1}\\
\textsuperscript{1}King's College London, \textsuperscript{2}Imperial College London\\
{\tt\small \{lin.3.li, haoyan.guan, michael.spratling\}@kcl.ac.uk, jianing.qiu17@imperial.ac.uk}
}

\begin{document}
\maketitle

\begin{abstract}

Large pre-trained Vision-Language Models (VLMs) like CLIP, despite having remarkable generalization ability, are highly vulnerable to adversarial examples. 
This work studies the adversarial robustness of VLMs from the novel perspective of the text prompt instead of the extensively studied model weights (frozen in this work). 
We first show that the effectiveness of both adversarial attack and defense are sensitive to the used text prompt.
Inspired by this, we propose a method to improve resilience to adversarial attacks by learning a robust text prompt for VLMs. 
The proposed method, named Adversarial Prompt Tuning (APT), is effective while being both computationally and data efficient.
Extensive experiments are conducted across 15 datasets and 4 data sparsity schemes (from 1-shot to full training data settings) to show APT's superiority over hand-engineered prompts and other state-of-the-art adaption methods. APT demonstrated excellent abilities in terms of the in-distribution performance and the generalization under input distribution shift and across datasets.
Surprisingly, by simply adding one learned word to the prompts, APT can significantly boost the accuracy and robustness ($\epsilon=4/255$) over the hand-engineered prompts by +13\% and +8.5\% on average respectively. 
The improvement further increases, in our most effective setting, to +26.4\% for accuracy and +16.7\% for robustness. Code is available at \url{https://github.com/TreeLLi/APT}.
\end{abstract}

\vspace{-5mm}

\section{Introduction}
Large pre-trained Vision-Language Models (VLMs) such as CLIP~\citep{radford_learning_2021}, ALIGN~\citep{jia_scaling_2021}, BLIP~\citep{li_blip_2022}, \etc have emerged as general-purpose (a.k.a. foundation) models \citep{bommasani_opportunities_2022}, fostering ecosystems across numerous sectors within the realm of artificial intelligence~\citep{bommasani_opportunities_2022,qiu2023large}. 
As more research and applications build upon these foundation models, any failures or vulnerabilities inherent in them can cause cascading impacts on the performance and reliability of the downstream tasks. 
A critical issue unveiled by the recent studies \citep{mao_understanding_2023,zhao_evaluating_2023,inkawhich_adversarial_2023,schlarmann_adversarial_2023} is that these VLMs, like vision models, are highly vulnerable to adversarial examples \citep{szegedy_intriguing_2014}. 
Their output can be manipulated by human-imperceptible perturbations to the image \citep{mao_understanding_2023,schlarmann_adversarial_2023}, posing substantial safety implications and thereby raising serious concerns about the reliability and security of these models.

\begin{figure}
    \centering
    \includegraphics[width=\linewidth]{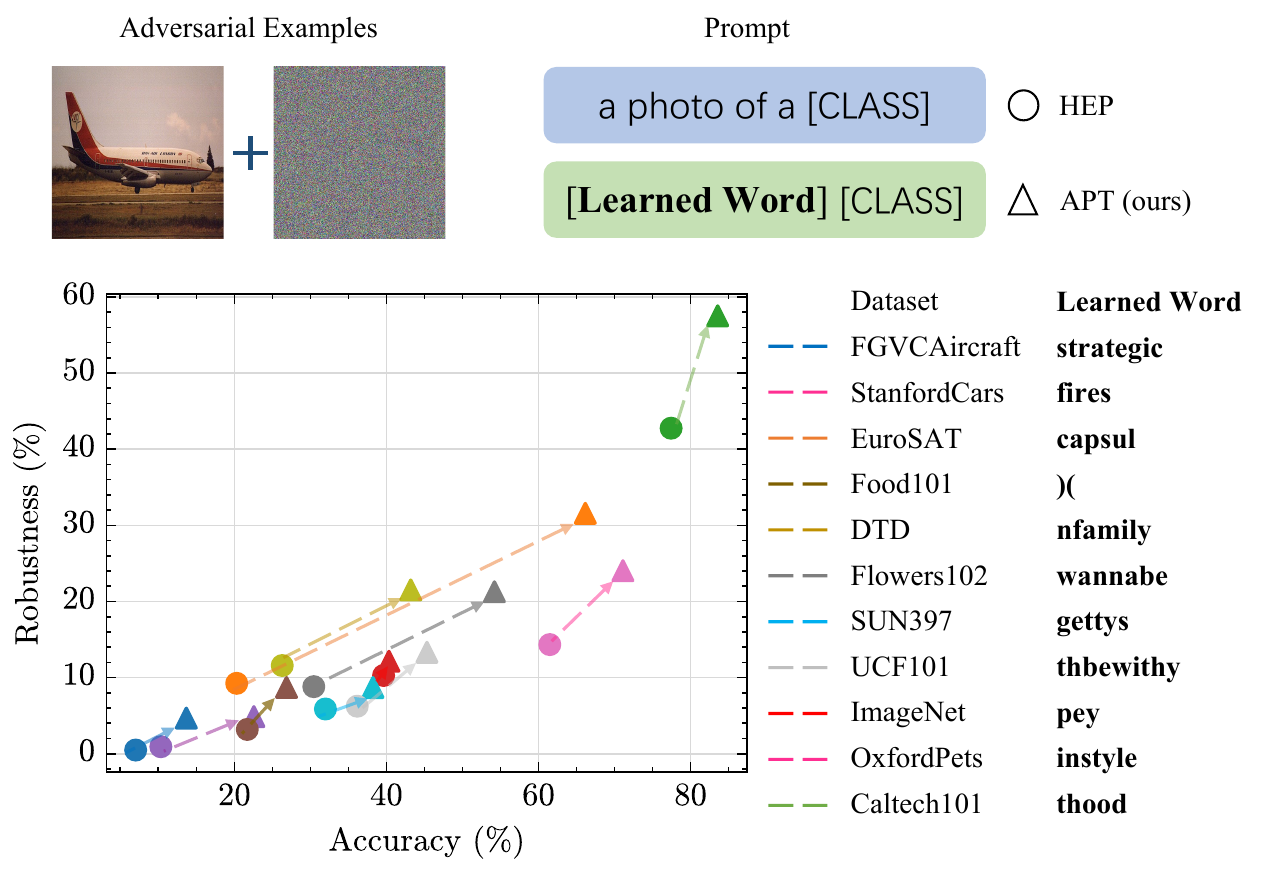}
    \vspace{-6mm}
    \caption{\textbf{Adding a learned ``word" to prompts boosts both accuracy and robustness ($\epsilon=4/255$) substantially over hand-engineered prompts (HEP) across 11 datasets}. 
    The dashed arrows indicate the performance boost.
    A ``word" is a learnable vector, which is interpreted in the last column of the figure.
    }
    \vspace{-4mm}
    \label{fig: one word boost}
\end{figure}

A prevalent paradigm \citep{bommasani_opportunities_2022} for the deployment of contemporary VLMs involves the initial pre-training of large models on large-scale datasets, followed by adapting for specific downstream tasks. 
Adaption is vital as it can often largely boost the performance for downstream tasks. 
A well-established approach to adaptation is fine-tuning the model weights \citep{yosinski_how_2014}.
However, fine-tuning all the model weights becomes prohibitively costly as pre-trained models scale up to tens or hundreds of billions of parameters, and can be even more unaffordable if adversarial training \citep{madry_towards_2018} is applied to improve adversarial robustness.  
Besides, fine-tuning may distort pre-trained features, and thus, hurt the out-of-distribution generalization performance \citep{kumar_fine-tuning_2022}. 
Therefore, parameter-efficient adaption methods \citep{ding_parameter-efficient_2023} that freeze all or most model weights become a promising solution. 

\begin{figure}
    \centering
    \includegraphics[width=.9\linewidth]{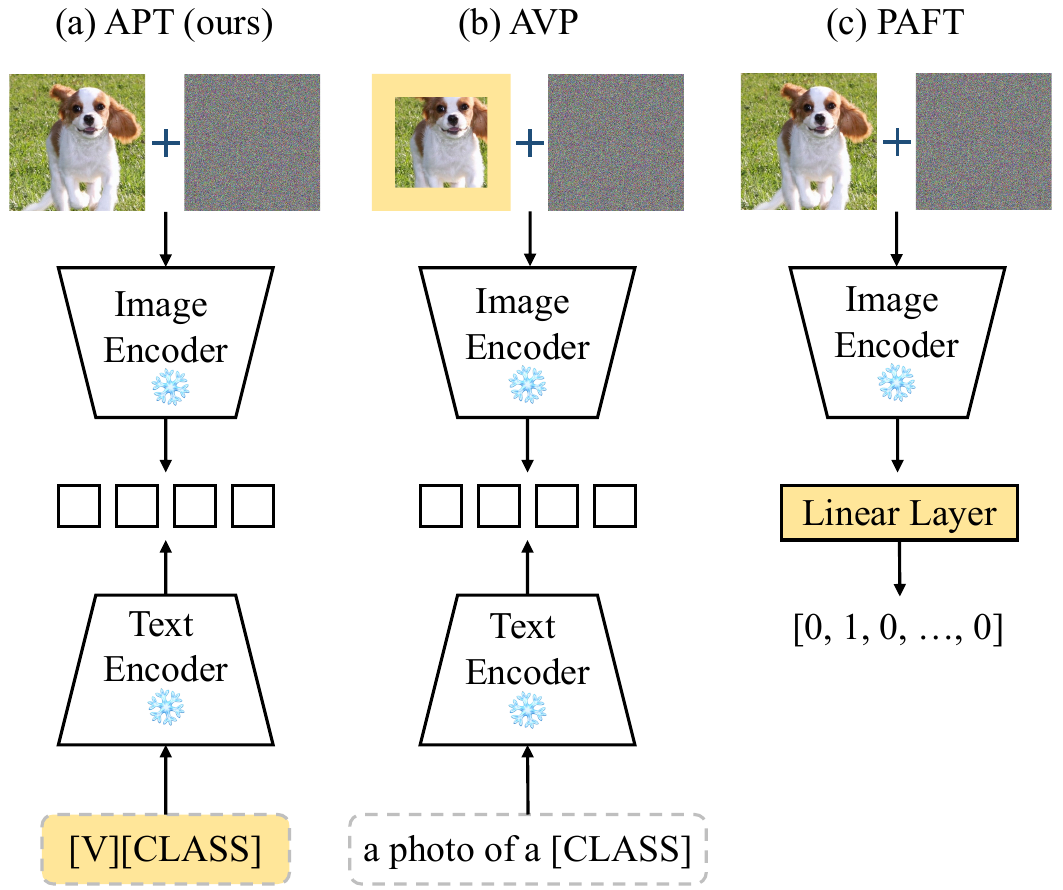}
    \vspace{-2mm}
    \caption{A high-level architectural comparison between our method Adversarial Prompt Tuning (APT), Adversarial Visual Prompting (AVP), and Partial Adversarial Fine-Tuning (PAFT). The learnable parameters are highlighted in yellow. Note that PAFT discards the entire text branch of CLIP.}
    \label{fig: architectural comparison among adaption methods}
    \vspace{-4mm}
\end{figure}

This work studies the problem of \textit{parameter-efficient adaption of pre-trained VLMs for adversarial robustness}. 
Current adaption methods for adversarial robustness focus on the model weights, \ie, adversarial fine-tuning~\citep{hendrycks_using_2019,chen_adversarial_2020,jiang_robust_2020,kim_adversarial_2020,luo_rethinking_2023} or image pixels, \ie, adversarial visual prompting \citep{chen_visual_2023,huang_improving_2023}. 
The text input to VLMs, a.k.a. prompt, has been rarely studied before for adversarial robustness despite its significant impact on the accuracy of VLMs \citep{zhou_learning_2022} and advantages such as naively supported by VLMs (so no need to modify architecture), parameter efficiency, \etc. 
This work aims to fill this gap by studying the effect of text prompt in adversarial robustness and proposing a new method to tune text prompt for improving adversarial robustness (see \cref{fig: architectural comparison among adaption methods}).
We focus on a category of VLMs resembling CLIP \citep{radford_learning_2021} as it represents a quintessential vision-language foundation model and has been used in many applications.

We start by investigating how the text prompt influences adversarial attack and defense on CLIP. Our key findings include: 1) the strength of adversarial attack is sensitive to the prompt used for generating adversarial examples;
2) the strongest adversarial examples are almost all generated when the prompt used for attack is the same as the prompt used by the victim model during inference;
3) the adversarial robustness of CLIP is sensitive to the prompt used for inference. 
The former two findings shed light on how to prompt for strong adversarial attack.


The last finding leads us to propose \textbf{A}dversarial \textbf{P}rompt \textbf{T}uning (APT) to learn robust text prompts for CLIP based on adversarial examples to improve its adversarial robustness. 
APT parameterizes prompts in the form of soft prompts~\citep{liu_pre-train_2021}, \ie, concatenating the class embedding with a sequence of learnable vectors (illustrated in \cref{fig: method overview}).
These vectors constitute the \textit{context} description of the data and class.
They can be unified to be shared by all classes or specific to each class.
Three different prompting strategies are then proposed to generate training adversarial examples on which the learnable vectors are optimized to minimize the predictive loss like CrossEntropy. 
Ultimately, the best prompting strategy we adopted is to generate training adversarial examples based on the latest updated prompts.


Extensive experiments are conducted to benchmark APT across 15 datasets and 4 data sparsity schemes, 1-, 4- and 16-shot learning and training with the entire training set.
APT is compared against the hand-engineered prompts proposed in CLIP \citep{radford_learning_2021} and the state-of-the-art adaption methods other than text prompting. APT is found to outperform these alternative methods in terms of the in-distribution performance and the generalization ability under distribution shift (the same classes yet different input distribution) and across datasets (different classes).
Three promising properties of APT are highlighted below:
\begin{itemize}
    \item Parameter-efficient: one prompt word is enough to boost performance substantially (see \cref{fig: one word boost}).
    \item Data-efficient: one shot is enough to boost performance considerably. 
    \item Effective: large performance boost and excellent trade-off between accuracy and robustness.
\end{itemize}
Overall, our work paves a new way for enhancing adversarial robustness for VLMs through text prompting.


\section{Related Works}

\textbf{Adapting pre-trained models for accuracy}.
In contrast to the traditional approach to fine-tune the entire model's parameters \citep{7426826}, parameter-efficient adaptation methods are investigated. Current parameter-efficient methods mainly contain three categories: prompt tuning \citep{zhou_learning_2022}, adapter tuning \citep{houlsby2019parameter,lin2020exploring} and linear probing \citep{kumar_fine-tuning_2022}. 
Prompt tuning modifies the input to adapt the model. 
According to the modality of the input, prompt tuning can be categorized into visual prompting \citep{zhou2023zegclip, bowman2023carte, yu2023prompting, zhang2022neural, loedeman2022prompt} for image input and text prompting \citep{ma2023understanding, park2023lanit, peng2023sgva, tao2023galip, lin2023being} and for text input. Adapter tuning inserts a small learnable module in the model to be trained for downstream tasks. 
Linear probing is performed by training only a linear layer attached to the end of the model.
In this paper, we investigate the application of text-driven prompt learning as a strategy for defending against adversarial attacks in the context of image recognition.

\textbf{Adversarial training}~\citep{goodfellow_explaining_2015} has been so far the most effective defense against adversarial examples \citep{athalye_obfuscated_2018}. 
It replaces the clean examples with adversarial examples generated on-the-fly during training. 
Adversarial training is well known to be expensive \citep{madry_towards_2018} and prone to overfitting \citep{wong_fast_2020,rice_overfitting_2020}. 
Numerous methods have been proposed to improve the efficiency \citep{shafahi_adversarial_2019,andriushchenko_understanding_2020,wong_fast_2020,jorge_make_2022,li_understanding_2023} and/or the effectiveness \citep{li_data_2023,wu_adversarial_2020,madry_towards_2018,zhang_theoretically_2019,wang_better_2023} of the algorithm.
However, most of them train models from scratch, while the adaption of pre-trained models for adversarial robustness is less studied. 
A line of works \citep{hendrycks_using_2019,chen_adversarial_2020,jiang_robust_2020,kim_adversarial_2020,luo_rethinking_2023} adapt pre-trained models for adversarial robustness by adversarial fine-tuning: fine-tuning model weights by adversarial training. 
According to the amount of parameters to be tuned, those methods are categorized as full adversarial fine-tuning and partial adversarial fine-tuning \citep{chen_adversarial_2020}.
Alternatively, \citet{chen_visual_2023} and \citet{huang_improving_2023} explore adversarial visual prompting \citep{bahng_exploring_2022} as a test-time defense to enhance adversarial robustness for pre-trained models. 
Our method aims at adapting pre-trained models by tuning text prompts, differing from the above works that adapt model weights or input images. More works on the adversarial robustness of VLMs are reviewed in \cref{app: related works}.


\section{Text Prompt for Adversarial Robustness}

\begin{figure}
    \centering
    \includegraphics[width=\linewidth]{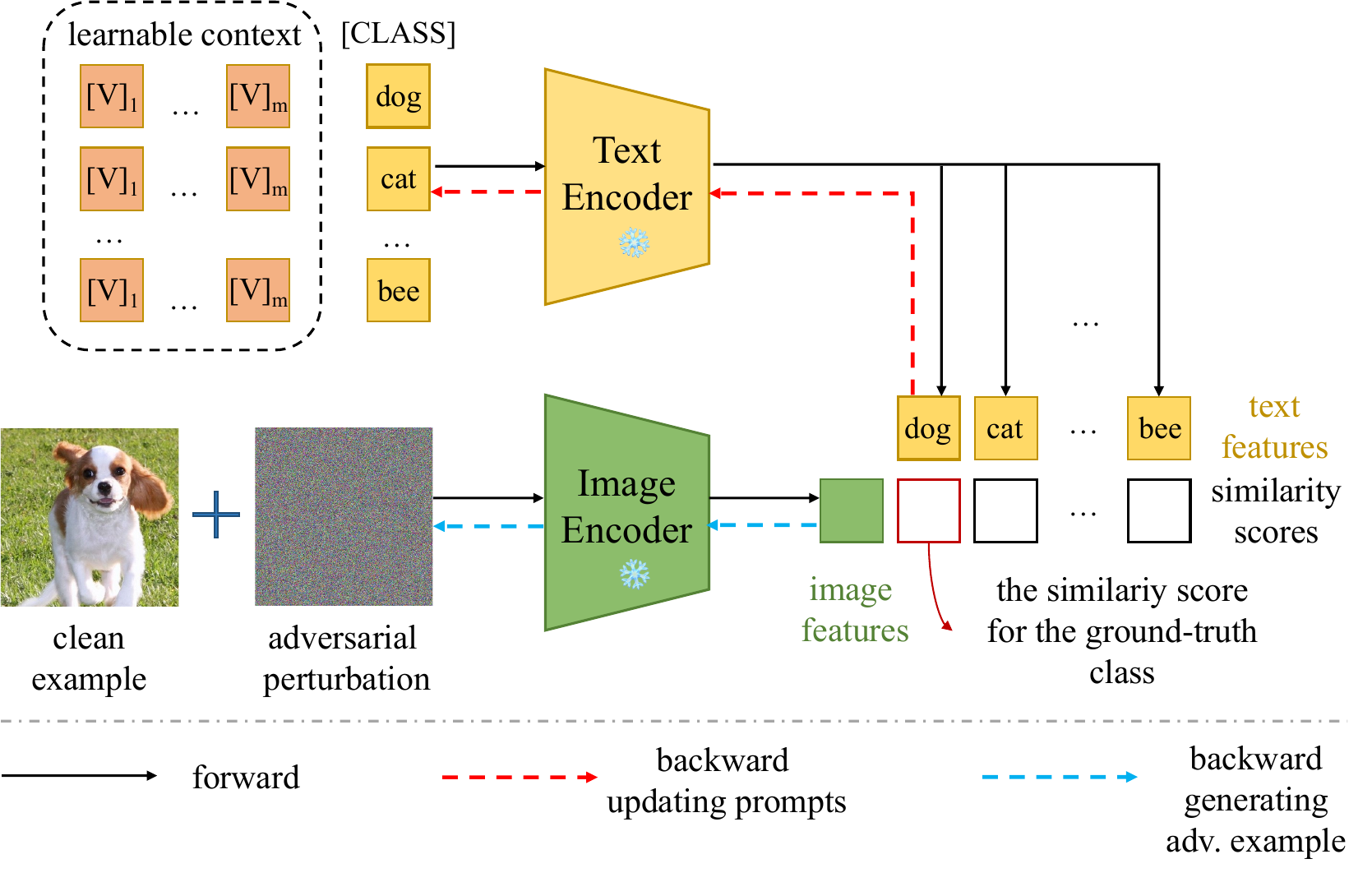}
    \vspace{-6mm}
    \caption{An overview of the proposed Adversarial Prompt Tuning (APT) method on CLIP-like VLMs. Both image and text encoders are frozen and only the prompt contexts are learnable. The learnable context can be unified for all classes or specific to each class. }
    \label{fig: method overview}
    \vspace{-4mm}
\end{figure}

\subsection{Review of CLIP}

As shown in \cref{fig: method overview}, CLIP consists of two primary components: an image encoder and a text encoder, parameterized by $\bm{\theta}_v$ and $\bm{\theta}_t$ respectively. 
They are used to extract the features from images and text respectively. 
Given an input image $\bm{x}_i$ and text $\bm{t}_j$, the respective features $\bm{z}_v^i$ and $\bm{z}_t^j$ are computed as:
\vspace{-2mm}
\begin{equation}
    \bm{z}_v^i = f(\bm{x}_i;\bm{\theta}_v), \ \ \ \ \bm{z}_t^j = f(\bm{t}_j; \bm{\theta}_t)
    \vspace{-2mm}
\end{equation}

A cosine similarity score is then calculated for each pair of image and text features to measure their alignment:
\vspace{-2mm}
\begin{equation}
    s_{i,j} = \cos(\bm{z}_v^i, \bm{z}_t^j)
    \label{equ: image text feature similarity}
    \vspace{-2mm}
\end{equation}
These similarity scores are analogous to the logit output of the classical vision model like ResNet \citep{he_deep_2016}.
The probability of $\bm{x}_i$ aligning with $\bm{t}_j$ is:
\vspace{-2mm}
\begin{equation}
    p_{i,j} = p(\bm{x}_i, \bm{t}_j) = \frac{\exp(s_{i,j})}{\sum_j\exp(s_{i,j})}
    \label{equ: inference probability}
    \vspace{-2mm}
\end{equation}

\begin{algorithm}
  \caption{Pseudo-code for $\ell_{\infty}$ adversarial attack on CLIP. Text is perturbed if $perturb\_t$ is true. $K$ is the step number. $\alpha$ ($\alpha'$) is the step size for perturbing image (text).}
  \label{algo: adv attack}
  \begin{algorithmic}[1]
    \Function{attack}{$\bm{x}$, $\bm{y}$, $\bm{t}$, $perturb\_t$}
      \State $\bm{\delta} = \text{uniform}(-\epsilon, \epsilon)$ \Comment{perturbation at image pixels}
      \State $\bm{\delta}' = \bm{0}$ \Comment{perturbation at word embeddings}
      \For{$1 \to K$}
            \State $\bm{x}' = \min(0, \max(\bm{x}+\bm{\delta}, 1))$
            \State $L = \mathcal{L}(\bm{x}', \bm{t}+\bm{\delta}', \bm{y};\bm{\theta_v}, \bm{\theta_t})$
            \State $\bm{\delta} = \min(-\epsilon, \max(\bm{\delta} + \alpha \cdot \Call{sign}{\nabla_{\bm{x}}L}, \epsilon))$
            \If{$perturb\_t$} \Comment{jointly perturb prompt}
                \State $\bm{\delta}' = \bm{\delta}' + \alpha' \cdot \nabla_{\bm{t}}L$
            \EndIf
      \EndFor
      \State \Return $\min(0, \max(\bm{x}+\bm{\delta}, 1))$
    \EndFunction
  \end{algorithmic}
\end{algorithm}

Two encoders are jointly pre-trained by maximizing the similarity scores for true image-text pairs, \ie, $i=j$ while minimizing the similarity scores for false pairs. 
Once pre-trained, CLIP can be applied to perform zero-shot image classification by using the text description of the classes in the target dataset as the text prompts and predicting the most probable class:
\vspace{-2mm}
\begin{equation}
    \text{arg} \max_j\ p_{i,j}
    \vspace{-2mm}
\end{equation}

By default, CLIP constructs the prompt for each class using a template of ``\texttt{a photo of a [CLASS]}" where \texttt{[CLASS]} is the name of a class. We define the content in the prompt other than \texttt{[CLASS]} as the \textit{context}. A prompt can be formulated as:
\vspace{-2mm}
\begin{equation}
    t_j = [\text{context}_{\text{front}}][\text{CLASS}_j][\text{context}_{\text{end}}]
    \label{equ: text prompt general formulation}
    \vspace{-2mm}
\end{equation}
Theoretically, the context can be arbitrary which provides a new dimension for adapting a frozen, pre-trained, VLM. 
Empirically, it has been shown that tuning the text prompt context can significantly impact the performance on the target dataset \citep{zhou_learning_2022,zhou_conditional_2022}.
Note that some specific details are ignored in the above review for simplicity. Please refer to the original work of CLIP \citep{radford_learning_2021} for the complete specification.



\subsection{The Sensitivity of Robustness to Prompts}
\label{sec: CLIP adversarial examples}

A common strategy \citep{mao_understanding_2023,zhao_evaluating_2023} to generate adversarial examples for VLMs is to search for a perturbation $\bm{\delta}_i$ for input $\bm{x}_i$ to maximize the (cosine) dissimilarity between the image feature, $\bm{z}_v^i$, and the text feature of the corresponding ground-truth class prompt, $\bm{z}_t^{y_i}$. Assuming $\bm{\delta}$ is bounded by the $\epsilon$-ball of the $p$-norm, it can be formulated as: 
\vspace{-2mm}
\begin{equation}
    \text{arg} \max_{\lVert \bm{\delta}_i\rVert_p\leq\epsilon} \mathcal{L}(\bm{x}_i+\bm{\delta}_i, \bm{t}', y_i;\bm{\theta_v}, \bm{\theta_t})
    \label{equ: adversarial generation}
    \vspace{-2mm}
\end{equation}
This differs from the conventional formulation \citep{li_understanding_2023} due to the presence of the text encoder, $\bm{\theta}_t$, and text prompt, $\bm{t}'$ (which can be different from the one used for inference, $\bm{t}$, in \cref{equ: inference probability}).
The effectiveness of adversarial examples generated by \cref{equ: adversarial generation} is dependent on the text encoder and text prompt since the gradients used for constructing adversarial examples are dependent (due to \cref{equ: image text feature similarity}) on the text features. 
Nevertheless, the influence of the text encoder is fixed and can be ignored as in this work its weights are frozen after pre-training. 
An implementation of the above attack algorithm is illustrated in \cref{algo: adv attack}.

\begin{figure}
    \centering
    \includegraphics[width=\linewidth]{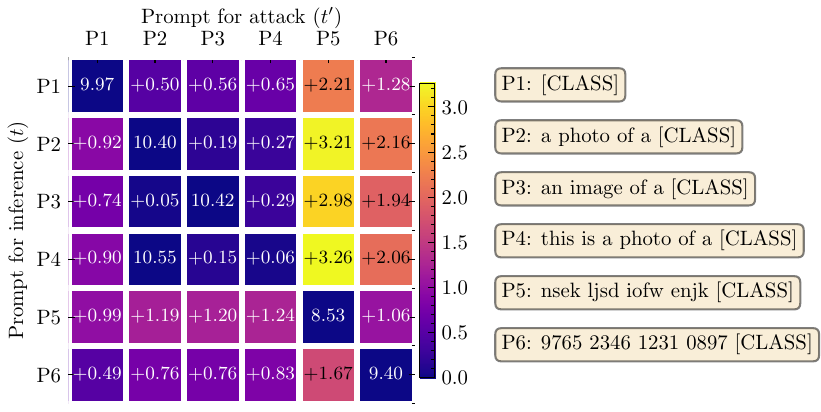}
    \vspace{-6mm}
    \caption{The robustness averaged over 11 datasets of pre-trained CLIP as varied prompts are used for inference, $\bm{t}$, (rows) and adversarial attack, $\bm{t}'$, (columns). The image encoder backbone is ViT-B/32. Robustness is evaluated against PGD100. Prompts 1 to 4 are manually constructed. Prompts 5 and 6 are randomly sampled from English characters and numbers respectively. For each row, the cell of the most malicious $\bm{t}'$, \ie, with the lowest robustness is annotated by the absolute robustness while the rest are annotated by the relative robustness, \ie, the amount exceeding the row minimum. Cells are colored according to the relative robustness.}
    \label{fig: robustness varied prompts}
\vspace{-4mm}
\end{figure}

Now the question is how $\bm{t}'$ should be selected to maximize the strength of the attack. 
A common choice \citep{zhao_evaluating_2023,mao_understanding_2023} is to use the same prompt as the one used for inference assuming that the attackers have access to this information, \ie a white-box threat model. 
To validate this, we fix the prompt for inference and vary the prompt for attack. 
It is observed (see \cref{fig: robustness varied prompts}) that the strength of the attack is sensitive to $\bm{t}'$. The robustness can vary a lot when different $\bm{t}'$ are used to attack the same $\bm{t}$.
Importantly, the lowest robustness is achieved when $\bm{t}' = \bm{t}$ in all cases except for the inference prompt P4.
Nevertheless, in that case, the gap between the robustness when using the attack prompt P4 (\ie the same inference and attack prompts) and the lowest robustness (produced by the attack prompt P2) is very small, 0.06\%. 
It is therefore vital for attackers to have access to the prompts used by the model users to construct strong attack.



Another intriguing observation in \cref{fig: robustness varied prompts} is that the (lowest) robustness varies with the prompt for inference. 
For instance, by simply changing the inference prompt from P5 (``\texttt{nsek ljsd iofw enjk [CLASS]}") to P4 (``\texttt{this is a photo of a [CLASS]}"), the worst-case robustness (evaluated by $\bm{t}'=\bm{t}$) increases from 8.53\% to 10.55\%.  

\section{Adversarial Prompt Tuning (APT)}

Motivated by the above observation, we hypothesize that the adversarial robustness of VLMs is sensitive to the text prompt used for inference, $\bm{t}$.
Therefore, we propose to improve the adversarial robustness of VLMs through adversarially tuning the prompt. 
Specifically, we aim at learning text prompt contexts that make the model more robust to adversarial attacks. 

\subsection{Prompt Parameterization}
\label{sec: context parameterization}
We first parameterize the context in a text prompt (\cref{equ: text prompt general formulation}) to be learnable. Following \citet{zhou_learning_2022}, the context of a class $C_j$ is formulated by a sequence of $M$ vectors, $[V]_{m,j}$ ($m\in{1,...,M}$), defined in the word embedding space rather than as raw text. This enables the parameters to be continuous for more flexibility compared to the discrete ones of the textual formulation. Each vector has the same dimension as a word embedding, \ie, 512 for CLIP. The final input to the text encoder is the concatenation of the context vectors and the word embedding of the class or the class embedding for short, $[C_j]$, as 
\vspace{-2mm}
\begin{equation}
    \bm{t}_j = [V]_{1,j}...[V]_{m,j}[C_j]
    \label{equ: text prompt learnable formulation}
    \vspace{-2mm}
\end{equation}
Theoretically, $[C_j]$ can be placed at an arbitrary position inside the sequence of context vectors. For simplicity, we only test three positions: \textit{front}, \textit{middle} and \textit{end}. Empirically, no distinction is observed among the results for these three positions (see \cref{sec: ablation study class embedding position}), so \textit{end} is used by default. 

Furthermore, we employ two variants of context parameterization: \textit{Unified Context} (UC) and \textit{Class-Specific Context} (CSC). 
In UC, the same context vectors are shared by all classes so there is only one sequence of context vectors no matter how many classes are used. 
In contrast, CSC assigns separate context vectors to each class so different classes are allowed to have different, tailored, contexts. 
The number of parameters for CSC increases linearly with the number of classes, $C$, in the dataset. 
Given the same context length, the CSC variant has $C$ times more parameters than the UC variant.
This may benefit learning complicated tasks but at the expense of requiring more training data to mitigate overfitting. A detailed empirical comparison is given in \cref{sec: uc vs csc}, but in summary, UC (CSC) is more effective when the training data is limited (abundant).

\subsection{Prompt Optimization}
\label{sec: method context optimization}
To improve adversarial robustness, we train the prompt contexts using adversarial training \citep{madry_towards_2018}:
\vspace{-2mm}
\begin{equation}
    \text{arg} \min_{\bm{t}} \mathbb{E}_{i\in B} \mathcal{L}(\bm{x}_i+\bm{\delta}_i, \bm{t}, y_i;\bm{\theta_v}, \bm{\theta_t})
    \label{equ: text prompt adv training}
    \vspace{-2mm}
\end{equation}
Where the perturbation $\bm{\delta}_i$ is generated on-the-fly by a training adversary as illustrated in \cref{algo: apt training}. 
Inside the prompt $\bm{t}$, only the context vectors have learnable parameters while the class embeddings are constant so optimizing the prompt is essentially optimizing the context vectors. 
Note that \cref{equ: text prompt adv training} can be easily extended to alternative adversarial training methods like TRADES \citep{zhang_theoretically_2019}.

\begin{algorithm}
  \caption{Pseudo-code of APT. $\bm{v}$ is the learnable context vectors. $\text{ATTACK}(\cdot)$ is defined in \cref{algo: adv attack}.}
  \label{algo: apt training}
  \begin{algorithmic}[1]
    \Function{train\_one\_iteration}{$\bm{x}$, $\bm{y}$}
        \State $\bm{t}=$\Call{g}{``\texttt{[CLASS]}"} \Comment{text to word embeddings}
        \State $\bm{t}=[\bm{v}, \bm{t}]$ \Comment{join context and class embedding}
            
        \If{\textit{constant}}
            \State $\bm{t}'=$\Call{g}{``\texttt{a photo of a [CLASS]}"}
            \State $\bm{x} =$ \Call{attack}{$\bm{x}$, $\bm{y}$, $\bm{t}'$, false}
        \ElsIf{\textit{on-the-fly}}
            \State $\bm{x} =$ \Call{attack}{$\bm{x}$, $\bm{y}$, $\bm{t}$, false}
        \ElsIf{\textit{perturbed}}
            \State $\bm{x} =$ \Call{attack}{$\bm{x}$, $\bm{y}$, $\bm{t}$, true}
        \EndIf
        
        \State $L = \mathcal{L}(\bm{x}, \bm{t}, \bm{y};\bm{\theta_v}, \bm{\theta_t})$
        \State $\bm{v} = \bm{v} - \ell \cdot \nabla_{\bm{v}}L$ \Comment{$\ell$ is learning rate}
    \EndFunction

  \end{algorithmic}
\end{algorithm}

The key design choice in \cref{equ: text prompt adv training} is the algorithm for generating $\bm{\delta}_i$. 
As discussed in \cref{sec: CLIP adversarial examples}, $\bm{\delta}_i$ is dependent on the text prompt $\bm{t}'$ used for attack that can be different from $\bm{t}$ in \cref{equ: text prompt adv training}. 
We propose three potential prompting strategies for generating training adversarial examples: \textit{constant}, \textit{on-the-fly} and \textit{perturbed} as formulated below respectively. 
\vspace{-2mm}
\begin{align}
    &\text{arg} \max_{\lVert \bm{\delta}_i\rVert_p\leq\epsilon} \mathcal{L}(\bm{x}_i+\bm{\delta}_i, \bm{t}^*, y_i;\bm{\theta_v}, \bm{\theta_t}) \label{equ: constant attack} \\
    &\text{arg} \max_{\lVert \bm{\delta}_i\rVert_p\leq\epsilon} \mathcal{L}(\bm{x}_i+\bm{\delta}_i, \bm{t}, y_i;\bm{\theta_v}, \bm{\theta_t}) \label{equ: updated attack} \\
    &\text{arg} \max_{\lVert \bm{\delta}_i\rVert_p\leq\epsilon, \bm{\delta}'} \mathcal{L}(\bm{x}_i+\bm{\delta}_i, \bm{t}+\bm{\delta}', y_i;\bm{\theta_v}, \bm{\theta_t}) \label{equ: perturbed attack}
    \vspace{-2mm}
\end{align}

The strategy \textit{constant} fixes the prompt for attack to a pre-defined one, ``\texttt{a photo of a [CLASS]}" in this case. 
The perturbation generated by this strategy for each image is constant during training regardless of the inference prompts since both model weights and attack prompts are fixed. 
This enables the reuse of adversarial image features and thus accelerates the prompt tuning process. 
However, it may not benefit or even hurt the performance as the perturbation now is no longer dynamically adversarial. 
In contrast, the strategy \textit{on-the-fly} generates adversarial examples based on the latest, updated, text prompts, $\bm{t}$ from \cref{equ: text prompt adv training}. 
This is the exact method used for adversarial evaluation, as discussed in \cref{sec: CLIP adversarial examples}.
Last, the strategy \textit{perturbed}, a.k.a. multimodal adversarial attack \citep{gan_large-scale_2020}, perturbs both images and text prompts (on top of the strategy \textit{on-the-fly}) to further enlarge the adversarial loss and hopefully to generate stronger adversarial examples.
This strategy was adopted before by \citet{gan_large-scale_2020} for adversarially training model weights. 
The algorithms for adversaries based on the above prompting strategies are illustrated in \cref{algo: adv attack}.

\textbf{Efficiency analysis}.
Analogous to adversarial training on model weights, the expense of APT is twofold: adversarial generation and prompt update. 
Supposing $K$ iterations of inner maximization, the cost of adversarial generation for the strategies \textit{constant} and \textit{on-the-fly} is $K$ times the number of forward and backward passes of the image encoder. 
The strategy \textit{perturbed} costs an extra $K$ forward and backward passes of the text encoder in addition to the expense of the strategy \textit{on-the-fly} for perturbing text prompts.
In practice, adversarial perturbation per image in the strategy \textit{constant}, once generated, can be cached and reused throughout training since the text prompt is constant. 
In contrast, adversarial perturbation has to be generated on-the-fly in each iteration for the other two strategies because the text prompt is variable.
The cost of prompt update is the same for all strategies: one forward pass of the image and text encoders plus one backward pass of the text encoder. 

A performance comparison among the above strategies is conducted in \cref{sec: ablation adversarial generation strategy}.
Empirically, the strategy \textit{on-the-fly} matches the effectiveness of strategy \textit{perturbed} while being much more effective than strategy \textit{constant}.
The strategy \textit{on-the-fly} is used by default as it achieves the best trade-off between effectiveness and efficiency.


\section{Experiments}

The experiments in this section were based on the following setup (more details in \cref{app: additional experiment setting}) unless otherwise specified. 
Following \citet{zhou_learning_2022}, 11 datasets were used to evaluate our method:  ImageNet \citep{deng2009imagenet}, Caltech101 \citep{fei2004learning}, OxfordPets \citep{parkhi2012cats}, StanfordCars \citep{krause20133d}, Flowers102 \citep{nilsback2008automated}, Food101 \citep{bossard2014food}, FGVCAircraft \citep{maji2013fine}, SUN397 \citep{xiao2010sun}, DTD \citep{cimpoi2014describing}, EuroSAT \citep{helber2019eurosat} and UCF101 \citep{soomro2012ucf101}.  
For each dataset, we evaluate with $N$-shots, meaning $N$ examples per class are randomly sampled from the entire training set for training. $N$ was either 1, 4, 16 or ``all", where the last means the entire training set was used.  
One exception was for ImageNet, where 100-shots was used instead of ``all'' because our computational resource is insufficient to run experiments on the full dataset.
All methods are evaluated on the entire test set regardless of the training data scheme used. 

\textbf{Models}. The default backbone for the image encoder is ViT-B/32 \citep{dosovitskiy2020image}.
The weights of image encoders were pre-trained using the state-of-the-art 
zero-shot adversarial robustness method TeCoA \citep{mao_understanding_2023}.
The necessity of robust pre-training is discussed in \cref{app: dependency on robust backbone}.

\textbf{Adversarial training and evaluation}. 
The PGD \citep{madry_towards_2018} attack is used for both training and evaluation. Two perturbation budgets, $\epsilon=1/255$ and $4/255$, are used following \citet{mao_understanding_2023} and \citet{croce_robustbench_2021} respectively. 
We use 3 steps with a step size of $2\epsilon/3$ for training and 100 steps with a step size of $\epsilon/4$ and random start for evaluation.
The inference prompts are used for attack as discussed in \cref{sec: CLIP adversarial examples}.

\textbf{Competitive methods}.
The proposed method is a text-prompting-based parameter-efficient adaption method so it is compared against two categories of related works: text prompting and parameter-efficient adaption methods. 
For text prompting, we compare our method against Hand-Engineered Prompts (HEP) which was originally proposed in CLIP and has been widely used subsequently \citep{zhou_learning_2022,zhou_conditional_2022,zhao_evaluating_2023,mao_understanding_2023}. 
The specific prompts used for each dataset are described in \cref{app: additional experiment setting}. For parameter-efficient adaption methods, we adopt Adversarial Visual Prompting (AVP) \citep{chen_visual_2023} and Partial Adversarial Fine-Tuning (PAFT) \citep{chen_adversarial_2020} for comparison. 
PAFT can be also viewed as the adversarial training variant of linear probing \citep{radford_learning_2021}.
A high-level architectural comparison between these adaption methods is shown in \cref{fig: architectural comparison among adaption methods}. 
The specification of AVP and PAFT is described in \cref{app: related works}.
All compared methods share the same frozen pre-trained image and text encoders. 

\subsection{In-Distribution Performance on 11 Datasets}

\label{sec: adv rob under different shots}
\begin{figure*}
    \centering
    \begin{subfigure}{\linewidth}
        \centering
        \includegraphics[width=.8\linewidth]{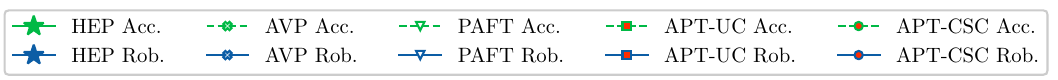}
    \end{subfigure}

    

    \hfill
    \begin{subfigure}{.24\linewidth}
        \includegraphics[width=\linewidth]{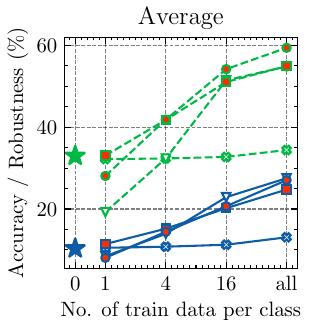}
    \end{subfigure}
    \hfill
    \begin{subfigure}{.24\linewidth}
        \includegraphics[width=\linewidth]{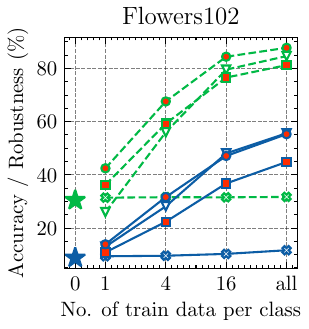}
    \end{subfigure}
    \hfill
    \begin{subfigure}{.24\linewidth}
        \includegraphics[width=\linewidth]{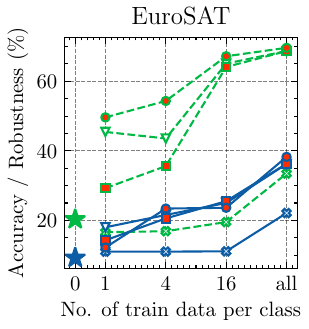}
    \end{subfigure}
    \hfill
    \begin{subfigure}{.24\linewidth}
        \includegraphics[width=\linewidth]{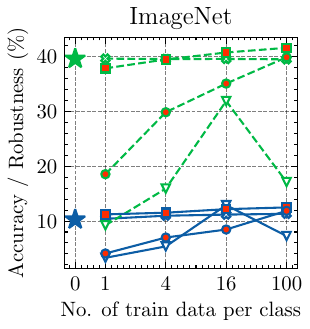}
    \end{subfigure}
    \hfill

    \hfill
    \begin{subfigure}{.24\linewidth}
        \includegraphics[width=\linewidth]{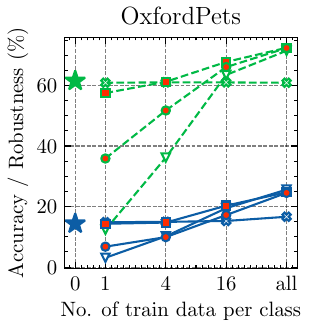}
    \end{subfigure}
    \hfill
    \begin{subfigure}{.24\linewidth}
        \includegraphics[width=\linewidth]{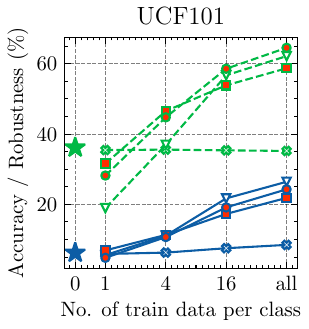}
    \end{subfigure}
    \hfill
    \begin{subfigure}{.24\linewidth}
        \includegraphics[width=\linewidth]{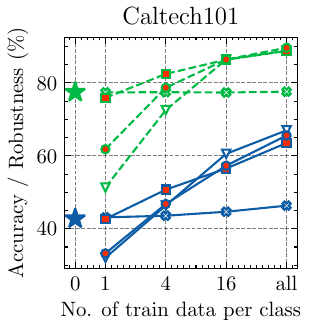}
    \end{subfigure}
    \hfill
    \begin{subfigure}{.24\linewidth}
        \includegraphics[width=\linewidth]{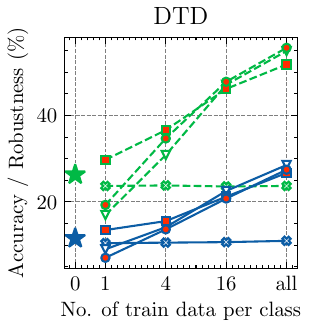}
    \end{subfigure}
    \hfill

    \hfill
    \begin{subfigure}{.24\linewidth}
        \includegraphics[width=\linewidth]{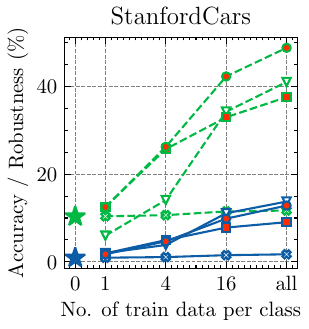}
    \end{subfigure}
    \hfill
    \begin{subfigure}{.24\linewidth}
        \includegraphics[width=\linewidth]{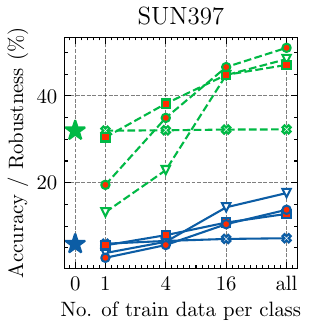}
    \end{subfigure}
    \hfill
    \begin{subfigure}{.24\linewidth}
        \includegraphics[width=\linewidth]{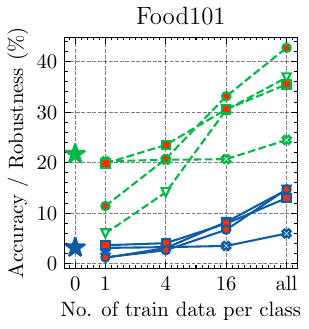}
    \end{subfigure}
    \hfill
    \begin{subfigure}{.24\linewidth}
        \includegraphics[width=\linewidth]{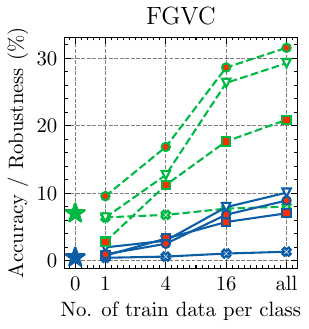}
    \end{subfigure}
    \hfill


    \vspace{-3mm}
    \caption{The in-distribution performance on 11 datasets and the averaged performance under different shots. $\epsilon=4/255$ and $M=16$.}
    \label{fig: shot results eps4}
    \vspace{-5mm}
\end{figure*}

This section benchmarks the proposed method and its competitors on the in-distribution performance, \ie, the training and test data are drawn from the (approximately) same distribution.
Specifically, models are adapted on the training set of a dataset and then evaluated on the test set of the same dataset. 
Below are the results for ViT-B/32 while the results for ResNet50 \citep{he_deep_2016} are given in \cref{app: resnet50}.


\begin{table}[]
\centering
\caption{The average performance for different $\epsilon$ and shots. The context length, $M$, for our methods is 16. The \textbf{best} and {\ul second best} results are highlighted under each metric. HEP are manually tuned on the target dataset so no strict control on the number of shots used. The results of HEP are copied under different shots in the table for the convenience of comparison.}
\vspace{-3mm}
\label{tab: average performance}
\resizebox{\columnwidth}{!}{%
\begin{tabular}{@{}llcccccccc@{}}
\toprule
\multicolumn{1}{c}{\multirow{2}{*}{$\epsilon$}} &
  \multicolumn{1}{c}{\multirow{2}{*}{Method}} &
  \multicolumn{2}{c}{1 shot} &
  \multicolumn{2}{c}{4 shots} &
  \multicolumn{2}{c}{16 shots} &
  \multicolumn{2}{c}{All} \\ \cmidrule(l){3-10} 
\multicolumn{1}{c}{} &
  \multicolumn{1}{c}{} &
  Acc. &
  Rob. &
  Acc. &
  Rob. &
  Acc. &
  Rob. &
  Acc. &
  Rob. \\ \midrule
\multirow{5}{*}{1/255} &
  HEP \citep{radford_learning_2021} &
  {\ul 45.2} &
  {\ul 32.1} &
  45.2 &
  32.1 &
  45.2 &
  32.1 &
  45.2 &
  32.1 \\ 
 &
  AVP \citep{chen_visual_2023} &
  44.6 &
  31.6 &
  45.0 &
  32.4 &
  45.7 &
  33.6 &
  50.6 &
  39.0 \\
 &
  PAFT \citep{chen_adversarial_2020} &
  30.6 &
  21.7 &
  46.9 &
  34.4 &
  66.4 &
  \textbf{51.0} &
  {\ul 71.1} &
  {\ul 56.9} \\ \cmidrule(l){2-10} 
 &
  APT-UC &
  \textbf{51.3} &
  \textbf{35.1} &
  \textbf{58.2} &
  \textbf{40.8} &
  {\ul 66.5} &
  49.0 &
  70.9 &
  54.3 \\
 &
  APT-CSC &
  39.9 &
  26.2 &
  {\ul 54.3} &
  {\ul 37.8} &
  \textbf{66.6} &
  {\ul 49.1} &
  \textbf{73.5} &
  \textbf{57.1} \\ \midrule
\multirow{5}{*}{4/255} &
  HEP \citep{radford_learning_2021} &
  {\ul 33.0} &
  10.3 &
  33.0 &
  10.3 &
  33.0 &
  10.3 &
  33.0 &
  10.3 \\ 
 &
  AVP \citep{chen_visual_2023} &
  32.2 &
  {\ul 10.5} &
  32.4 &
  10.8 &
  32.7 &
  11.3 &
  34.4 &
  13.1 \\
 &
  PAFT \citep{chen_adversarial_2020} &
  19.2 &
  8.5 &
  32.4 &
  13.9 &
  {\ul 51.5} &
  \textbf{22.9} &
  {\ul 54.9} &
  \textbf{27.5} \\ \cmidrule(l){2-10} 
 &
  APT-UC &
  \textbf{33.1} &
  \textbf{11.4} &
  {\ul 41.8} &
  \textbf{15.2} &
  51.1 &
  20.2 &
  {\ul 54.9} &
  24.8 \\
 &
  APT-CSC &
  28.1 &
  8.1 &
  \textbf{41.9} &
  {\ul 14.4} &
  \textbf{54.2} &
  {\ul 20.7} &
  \textbf{59.4} &
  {\ul 27.0} \\ \bottomrule
\end{tabular}%
}
\vspace{-3mm}
\end{table}

\textbf{Learned prompts vs. hand-engineered prompts}.
A comparison of different text prompting methods on the performance averaged over 11 datasets for various perturbation budgets, $\epsilon$, and shots is shown in \cref{tab: average performance}. Our method yields substantial improvement over HEP. 
Even for 1 shot, our method (UC variant) effectively boosts the accuracy and robustness over HEP and the improvement is remarkable for $\epsilon=1/255$, \ie, +6.1\% and +3.0\% for accuracy and robustness respectively. 
Furthermore, such improvement consistently increases with the number of shots. We highlight that when the entire training dataset is used, our method (CSC variant) achieves a substantial boost over HEP by +28.3\% (+26.4\%) and +25.0\% (+16.7\%) for accuracy and robustness respectively when $\epsilon=1/255$ (4/255).

For each specific dataset, our method shows improvement on all of them but the margin varies considerably. \cref{fig: shot results eps4} (\cref{fig: shot results eps 1} in Appendix) depicts the results for $\epsilon=4/255$ ($1/255$). The improvement is huge on some datasets such as Flowers102, EuroSAT, \etc, but relatively small on ImageNet. The reason why the result of ImageNet is small is because the model weights have been pre-trained with HEP on the entire training set of ImageNet using TeCoA~\citep{mao_understanding_2023} so HEP is supposed to be optimal in this setting. It is, therefore, promising that our method in this setting can still improve on this by an evident margin of, \eg, +1.1\% and +1.9\% for accuracy and robustness respectively (UC variant) when trained with 16 shots.

\textbf{APT vs. AVP and PAFT}.
A comparison of different types of adaption methods on the overall performance averaged over 11 datasets is given in \cref{tab: average performance}. It is observed that our method substantially outperforms AVP and PAFT in terms of both accuracy and robustness for 1 and 4 shots, suggesting that our method is more data-efficient than these alternatives.
Noticeably, PAFT is much inferior to our method and even considerably underperforms the baseline HEP method in the 1-shot setting. 
Furthermore, as more data is used for training, \ie, 16 and all shots, the superiority of our method compared to AVP in terms of both accuracy and robustness becomes more significant suggesting our method is much more effective than AVP in leveraging more data.
Meanwhile, compared to PAFT when using 16 and all shots, our method achieves a comparable robustness and a considerably higher accuracy, suggesting a much better trade-off between accuracy and robustness.

\label{sec: instability of paft}
For performance on each individual dataset as shown in \cref{fig: shot results eps4} (and \cref{fig: shot results eps 1} in the Appendix), we highlight that our methods exhibit a substantial improvement over PAFT regarding both accuracy and robustness on ImageNet when trained with 100 shots. 
We observe that PAFT suffered from underfitting on ImageNet with 100 shots for both $\epsilon$ settings. 
We tried training with more epochs (increased to 50 from 20 epochs) but found no effect.
This issue is even severer when a logistic classifier is applied as originally done for linear probing (\ie non-adversarial variant of PAFT) in CLIP \citep{radford_learning_2021}.
Note that the linear probing is also observed by \citet{zhou_learning_2022} to perform worse than zero-shot CLIP on ImageNet.


\label{sec: uc vs csc}

\textbf{Unified context vs. class-specific context}.
Two variants of our method (\cref{sec: context parameterization}) are compared.
The UC variant of our method in general outperforms the CSC variant when the training data is limited, \ie, 1 and 4 shots in \cref{tab: average performance}, but underperforms when the training data is relatively abundant, \ie, 16 and all shots. 
This is because the CSC variant has more parameters, and thus, larger capacity than the UC variant to learn from relatively larger-scale data. Nevertheless, the CSC variant also requires more data to mitigate overfitting due to the larger capacity. 
This likely accounts for the relatively poor performance of the CSC variant in 1- and 4-shot settings.
The above trends hold for most of the evaluated datasets, but it is also observed that for some datasets one variant is consistently superior to the other, as shown in \cref{fig: shot results eps4} (and Appendix \cref{fig: shot results eps 1}). 
For instance, the CSC variant achieves higher (lower) performance than the UC variant across all four data schemes on Flowers102 (ImageNet).

\subsection{Generalization of Learned Prompt Contexts}
\begin{table*}[]
\centering
\caption{The generalization of the prompts learned by our method on ImageNet to (1) datasets with input  distribution shifts and (2) the other 10 datasets. ``N/A" denotes that the corresponding method by design cannot generalize to the target setting. The results for both variants of our method are reported for the checkpoints trained with $M=4$ and 16 shots. $\epsilon=4/255$.}
\vspace{-3mm}
\label{tab: dist shift}
\begin{tabular}{@{}lcccccccccccc@{}}
\toprule
\multicolumn{1}{c}{\multirow{3}{*}{Method}} &
  \multicolumn{2}{c}{Source} &
  \multicolumn{8}{c}{Distribution Shifts} &
  \multicolumn{2}{c}{Cross Datasets} \\ \cmidrule(l){2-13} 
\multicolumn{1}{c}{} &
  \multicolumn{2}{c}{ImageNet} &
  \multicolumn{2}{c}{ImageNet-V2} &
  \multicolumn{2}{c}{ImageNet-Sketch} &
  \multicolumn{2}{c}{ImageNet-R} &
  \multicolumn{2}{c}{ObjectNet} &
  \multicolumn{2}{c}{10 Datasets Avg.} \\ \cmidrule(l){2-13} 
\multicolumn{1}{c}{} &
  Acc. &
  Rob. &
  Acc. &
  Rob. &
  Acc. &
  Rob. &
  Acc. &
  Rob. &
  Acc. &
  Rob. &
  Acc. &
  Rob. \\ \midrule
HEP &
  {\ul 39.86} &
  10.28 &
  {\ul 32.74} &
  7.49 &
  {\ul 17.40} &
  7.21 &
  21.46 &
  5.80 &
  {\ul 9.16} &
  1.15 &
  \textbf{32.32} &
  10.34 \\ 
AVP &
  39.61 &
  11.18 &
  32.68 &
  8.12 &
  17.39 &
  {\ul 7.69} &
  {\ul 21.47} &
  {\ul 6.25} &
  9.08 &
  {\ul 1.31} &
  31.37 &
  {\ul 11.13} \\
PAFT &
  31.92 &
  \textbf{12.90} &
  25.55 &
  \textbf{9.52} &
  10.02 &
  5.05 &
  13.34 &
  4.55 &
  5.57 &
  0.86 &
  N/A &
  N/A \\ \midrule
APT-CSC &
  37.18 &
  9.49 &
  28.93 &
  6.65 &
  12.72 &
  4.83 &
  15.06 &
  3.57 &
  7.17 &
  0.61 &
  N/A &
  N/A \\
APT-UC &
  \textbf{40.80} &
  {\ul 12.33} &
  \textbf{33.20} &
  {\ul 9.04} &
  \textbf{18.35} &
  \textbf{8.04} &
  \textbf{22.66} &
  \textbf{6.97} &
  \textbf{9.31} &
  \textbf{1.49} &
  {\ul 32.01} &
  \textbf{11.84} \\ \bottomrule
\end{tabular}
\vspace{-4mm}
\end{table*}

This section assesses the generalization ability beyond in-distribution performance of the text prompts learned by our method. Two types of generalization are evaluated: distribution shift and cross dataset. For both tests, we use ImageNet as the source dataset on which the model was adapted using the methods to be tested. We then evaluate the performance of these ImageNet-adapted models on the target datasets with the same classes yet different data distributions (\ie the distribution shift test) and the target datasets with different classes (\ie the cross dataset test). Specifically, following the work of \citet{li2023oodrobustbench}, we use four ImageNet variant datasets, ImageNet-V2 \citep{recht_imagenet_2019}, ImageNet-Sketch, ImageNet-R \citep{hendrycks_many_2021} and ObjectNet \citep{barbu_objectnet_2019}, to represent different kinds of distribution shift. We use the remaining ten of the original eleven datasets for the cross dataset test.

Our method exhibits the best overall generalization ability on both tests among all compared methods (\cref{tab: dist shift}). 
It achieves the highest accuracy and robustness on most target datasets. It is also noteworthy that PAFT, despite having the best robustness on the source dataset ImageNet, performs poorly under most distribution shifts, \eg, its relative robustness improvement over ours (UC) drops from +0.57\% on ImageNet to -2.99\% on ImageNet-Sketch and -2.42\% on ImageNet-R. 
More importantly, PAFT, unlike AVP and our method, cannot deal with the novel classes that were unseen during training due to its rigid, hard-coded, linear layer. Hence, it is not applicable to the cross dataset test. The same issue also applies to the CSC variant of our method.


\subsection{Trade-off Between Accuracy and Robustness}
\vspace{-1mm}

\begin{figure}
    \centering
    \includegraphics[width=\linewidth]{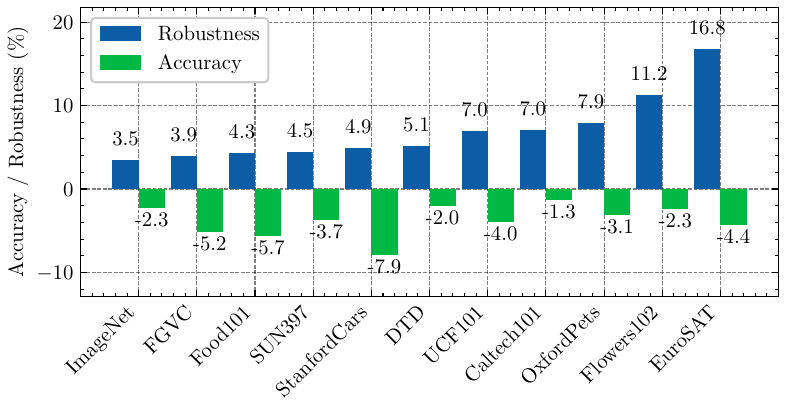}
    \vspace{-8mm}
    \caption{The performance improvement per dataset of our adversarially-trained prompt over the standardly-trained prompt for unified context ($M=16$). The results are reported on the checkpoints trained on 16 shots. $\epsilon=4/255$.}
    \label{fig: comparison against CoOp}
    \vspace{-6mm}
\end{figure}

As shown in \cref{fig: comparison against CoOp}, we compare our adversarially-trained prompt contexts to the standardly-trained prompt contexts \citep{zhou_learning_2022} based on the unified context. 
In general, the adversarially-trained prompts improve robustness at the expense of accuracy when compared to the standardly-trained prompts. 
This trade-off between accuracy and robustness is somewhat expected as it also happens to adversarially-trained vision models \citep{tsipras_robustness_2019}. 
More importantly, we found that for most datasets the improvement in robustness surpasses the reduction in accuracy. 
Taking an example of Flowers102, a significant improvement of +11.2\% in robustness is achieved with a sacrifice in accuracy of merely -2.3\%. This observation suggests an attractive trade-off between accuracy and robustness of our method.

\subsection{Reliability of Adversarial Evaluation}
\vspace{-1mm}
\label{sec: adv eval verification}
To verify that our evaluation of adversarial robustness is reliable, we first additionally evaluate the adversarial robustness of our methods using a diverse set of attacks including TPGD \citep{zhang_theoretically_2019}, CW \citep{carlini_towards_2017} and AutoAttack \citep{croce_reliable_2020}.
Next, to further exclude the possibility of our method masking the gradients for the particular prompts \citep{athalye_obfuscated_2018}, we evaluate our methods using the adversarial examples transferred from other prompts with the same model as defined in \cref{fig: robustness varied prompts}.
The results are described in \cref{sec: verification on adversarial evaluation}.
In summary, the robustness advantage of our methods is not a consequence of overfitting to the particular attack or obfuscated gradients \citep{athalye_obfuscated_2018}

\subsection{Ablation Study}
\vspace{-1mm}
To dissect the design choice of our method, we conduct ablation study on the length of prompt context, $M$, in \cref{sec: prompt context length}, the position of class embedding in \cref{sec: ablation study class embedding position}, and the prompting strategy for training adversarial generation in \cref{sec: ablation adversarial generation strategy}. 

\vspace{-1mm}
\section{Limitation}
\vspace{-1mm}
\label{sec: limitation}

APT has two limitations.
First, it is challenging to interpret the learned context vectors.
The semantics of the learned context, when decoded by the nearest words, appears to be irrelevant to the data and sometimes even uninterpretable.
Second, the effectiveness of APT depends on the pre-trained model weights.
We observe that APT and other parameter-efficient adaption methods including AVP and PAFT cannot effectively boost adversarial robustness for the standardly-pre-trained overly vulnerable model.
This is somewhat reasonable because the number of tunable parameters is dramatically limited compared to those of the image and text encoders esp. for the UC variant of our method.
We discuss the above two issues in detail in \cref{app: interpreting learned contexts,app: dependency on robust backbone}.

\vspace{-1mm}
\section{Conclusion}
\vspace{-1mm}
This work studies the adversarial robustness of VLMs from the novel perspective of the text prompt. 
We first demonstrate that both adversarial attack and defense for VLMs are sensitive to the used text prompt.
We then propose Adversarial Prompt Tuning (APT) to learn robust text prompts for CLIP based on adversarial examples to improve its adversarial robustness.
Extensive experiments are conducted to demonstrate the effectiveness of APT for both the in-distribution performance and the generalization ability under distribution shift and across datasets.
APT is also parameter- and data-efficient.
Given the promising performance of APT, our work paves a new way for enhancing adversarial robustness for VLMs through text prompting.

{
    \small
    \bibliographystyle{ieeenat_fullname}
    \bibliography{reference,references}
}

\clearpage
\appendix
\newpage

\section{Related works}
\label{app: related works}
\subsection{Adversarial Defense for VLMs}

Research on adversarial defense has concentrated on vision models, and few previous works consider VLMs \citep{gan_large-scale_2020,yang_defending_2021,mao_understanding_2023}.
\citet{gan_large-scale_2020} first employ adversarial training to train VLMs from scratch for improved clean performance. 
\citet{yang_defending_2021} equip standard VLMs with an auxiliary network and a robust feature fusion layer and performs adversarial training on these new modules. 
\citet{mao_understanding_2023} uses full adversarial fine-tuning to adapt a pre-trained CLIP model for zero-shot adversarial robustness. 
All of the above works study the adversarial robustness of VLMs through the lens of model weights, which leaves unexplored the question of how the text prompt affect adversarial robustness.
To fill this blank, we analyze the sensitivity of adversarial robustness to text prompts and propose a new method to optimize the text prompt to increase adversarial robustness.

\subsection{Adversarial Visual Prompting}
\label{app: related works avp}
AVP \citep{chen_visual_2023} combines visual prompting \citep{bahng_exploring_2022} with adversarial training to counteract adversarial perturbations. The visual prompt perturbation, parameterized by $\bm{\phi}$, is applied to the input image, $\bm{x}$, so that the prompted image is given by $\bm{x}_{vp} = \bm{x} + \bm{\phi}$. AVP optimizes visual prompt $\bm{\phi}$ to jointly minimize both clean and adversarial losses:
\begin{equation}
    \text{arg} \min_{\bm{\phi}} \mathcal{L}(\bm{x}+\bm{\phi}, \bm{t}, y;\bm{\theta_v}, \bm{\theta_t}) + \mathcal{L}(\bm{x}+\bm{\delta}+\bm{\phi}, \bm{t}, y;\bm{\theta_v}, \bm{\theta_t})
\end{equation}
where $\bm{\delta}$ is generated by \cref{algo: adv attack} independent of $\bm{\phi}$. 

\subsection{Partial Adversarial Fine-Tuning}
\label{app: related works paft}
PAFT \citep{chen_adversarial_2020} discards the text encoder branch of CLIP and attaches an extra linear layer, parameterized by $\bm{\theta_l}$, on top of the frozen image encoder, $\bm{\theta_{v}}$, to form a new classifier.
The output of the linear classifier has the same dimension as the number of classes. 
PAFT optimizes $\bm{\theta_l}$ to minimize the adversarial loss:
\begin{equation}
    \text{arg} \min_{\bm{\theta_l}} \mathcal{L}(\bm{x}+\bm{\delta}, y;\bm{\theta_v}, \bm{\theta_l})    
\end{equation}
where $\bm{\delta}$ is generated using the conventional PGD attack on the new classifier.

\section{Experiment Setting}
\label{app: additional experiment setting}

\textbf{Data}.
The 11 datasets were selected to constitute a comprehensive benchmark, encompassing a diverse array of vision tasks including generic objects classification, scene recognition, action classification, fine-grained classification, textures recognition and satellite imagery analysis.
They were split into training and test sets in the same way as \citet{zhou_learning_2022}. 
The $N$-shot images, once sampled, are fixed so that all evaluated methods are trained on the exactly same data for a fair comparison.

\textbf{Model}.
The text encoder is the default pre-trained model from CLIP \citep{radford_learning_2021}. 
For both image and text branches, we adopt the same data pre-processing as CLIP \citep{radford_learning_2021}.

\subsection{General Training Set-ups}
\textbf{APT}
was trained by Stochastic Gradient Descent (SGD) using a cosine learning rate schedule with an initial learning rate of 0.002. The batch size was 32. 
For ImageNet, the number of epochs was 20, 20, 50 and 20 for 1, 4, 16 and all shots respectively, and for other datasets was 50, 100, 200, 200.
The number of epochs used with ImageNet was reduced due to limited computational resources. 
Results were reported for the last checkpoint.

\textbf{AVP}
was implemented in the mode of padding with a prompt size of 52 to match the number of parameters of our method for the unified context. 
Although a class-specific variant of AVP was proposed in \citep{chen_visual_2023} and was observed to outperform its unified variant on CIFAR10 with ResNet18, we found  that the class-specific variant of AVP generalized poorly to our experimental set-ups, \ie, CLIP plus 11 diverse datasets. We therefore decided to use the unified variant of AVP.
Following \citep{chen_visual_2023}, a cosine learning rate schedule with an initial learning rate of 0.1 was used. 
The number of epochs, corresponding to 1/4/16/all shots, was 20/50/100/100 for non-ImageNet datasets and 20/20/50/20 for ImageNet. 
The implementation was based on the open-source code of \citet{chen_visual_2023}.

\textbf{PAFT}
was trained, following \citet{chen_adversarial_2020}, for 20/50/100/100 epochs with an initial learning rate of 0.1 decayed by 0.1 at epochs 5/15/30/30 and 10/25/50/50 for 1/4/16/all shots,  respectively, for non-ImageNet datasets. 
For ImageNet, the number of training epochs was 20/20/50/20 for 1/4/16/100 shots respectively.

\begin{table}[]
\centering
\caption{Datasets statistics and their prompts}
\label{tab: dataset}
\resizebox{\columnwidth}{!}{%
\begin{tabular}{@{}lcl@{}}
\toprule
\multicolumn{1}{c}{Dataset} & Classes & \multicolumn{1}{c}{Hand-Engineered Prompt} \\ \midrule
ImageNet   & 1000 &   a photo of a [CLASS]     \\
Caltech101        & 100 &    a photo of a [CLASS]      \\
OxfordPets & 37 & a photo of a [CLASS], a type of pet  \\ 
StanfordCars & 196 & a photo of a [CLASS] \\ 
Flowers102 & 102 & a photo of a [CLASS], a type of flower \\ 
Food101 & 101 & a photo of a [CLASS], a type of food \\ 
FGVCAircraft & 100 & a photo of a [CLASS], a type of aircraft \\ 
SUN397 & 397 & a photo of a [CLASS] \\ 
DTD & 47 & [CLASS] texture \\ 
EuroSAT & 10 & a centered satellite photo of a [CLASS] \\ 
UCF101 & 101 & a photo of a person doing [CLASS] \\ \bottomrule
ImageNetV2 & 1000 & a photo of a [CLASS] \\
ImageNet-Sketch & 1000 & a photo of a [CLASS] \\
ImageNet-A & 200 & a photo of a [CLASS] \\
ImageNet-R & 200 & a photo of a [CLASS] \\ \bottomrule
\end{tabular}%
}
\end{table}

\section{Additional Results}

\subsection{In-Distribution Performance for $\epsilon=1/255$}

\begin{figure*}
    \centering
    \begin{subfigure}{\linewidth}
        \centering
        \includegraphics[width=\linewidth]{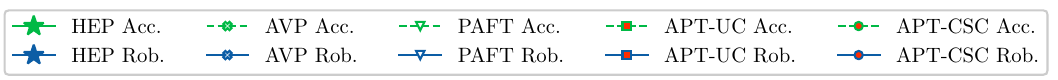}
    \end{subfigure}
    


    
    \hfill
    \begin{subfigure}{.27\linewidth}
        \includegraphics[width=\linewidth]{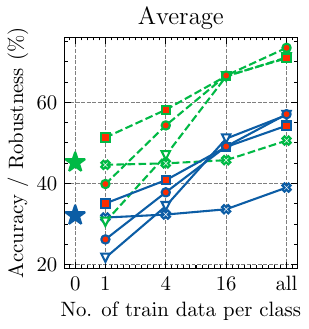}
    \end{subfigure}
    \hfill
    \begin{subfigure}{.27\linewidth}
        \includegraphics[width=\linewidth]{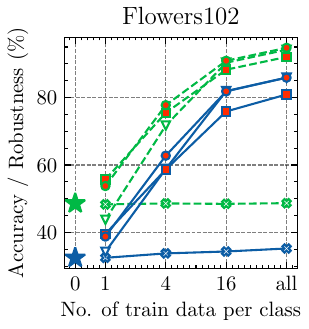}
    \end{subfigure}
    \hfill
    \begin{subfigure}{.27\linewidth}
        \includegraphics[width=\linewidth]{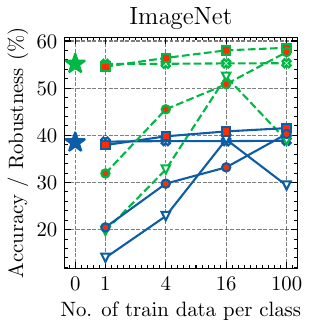}
    \end{subfigure}
    \hfill

    \hfill
    \begin{subfigure}{.27\linewidth}
        \includegraphics[width=\linewidth]{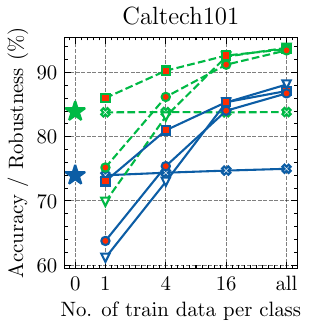}
    \end{subfigure}
    \hfill
    \begin{subfigure}{.27\linewidth}
        \includegraphics[width=\linewidth]{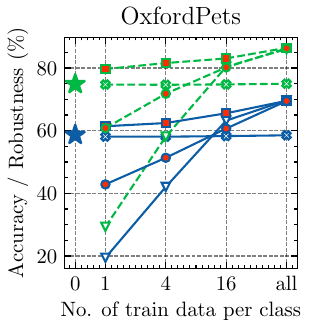}
    \end{subfigure}
    \hfill
    \begin{subfigure}{.27\linewidth}
        \includegraphics[width=\linewidth]{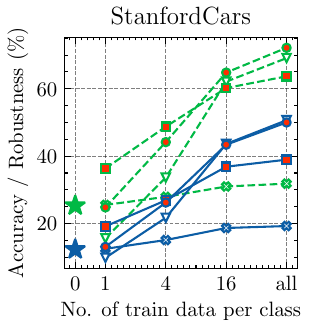}
    \end{subfigure}
    \hfill

    \hfill
    \begin{subfigure}{.27\linewidth}
        \includegraphics[width=\linewidth]{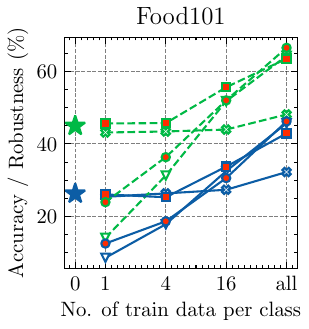}
    \end{subfigure}
    \hfill
    \begin{subfigure}{.27\linewidth}
        \includegraphics[width=\linewidth]{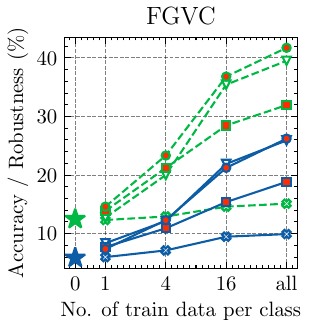}
    \end{subfigure}
    \hfill
    \begin{subfigure}{.27\linewidth}
        \includegraphics[width=\linewidth]{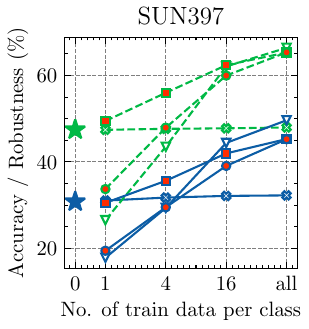}
    \end{subfigure}
    \hfill

    \hfill
    \begin{subfigure}{.27\linewidth}
        \includegraphics[width=\linewidth]{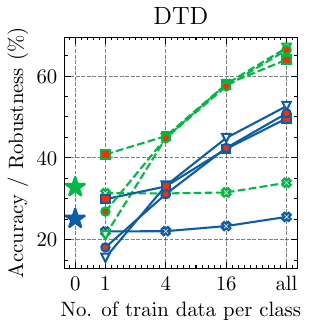}
    \end{subfigure}
    \hfill
    \begin{subfigure}{.27\linewidth}
        \includegraphics[width=\linewidth]{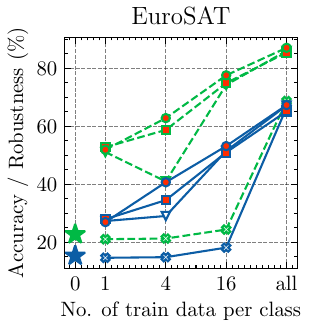}
    \end{subfigure}
    \hfill
    \begin{subfigure}{.27\linewidth}
        \includegraphics[width=\linewidth]{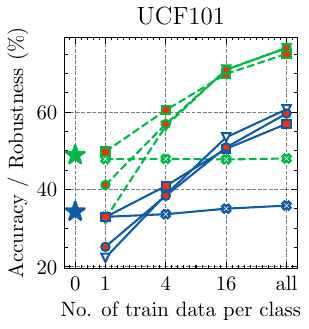}
    \end{subfigure}
    \hfill

    \caption{The in-distribution performance on 11 datasets and the averaged performance under different shots. $\epsilon=1/255$ and M = 16.}
    \label{fig: shot results eps 1}
    
\end{figure*}

\cref{fig: shot results eps 1} depicts the results for $\epsilon=1/255$ on each dataset. 
The trends are generally consistent to those observed for $\epsilon=4/255$ in \cref{fig: shot results eps4}.

\subsection{Generalization to Other Architectures}
\label{app: resnet50}
\begin{table}[]
\centering
\caption{The average performance for ResNet50 with $\epsilon=1/255$. $M=16$ and16 shots.}
\label{tab: rn50}
\begin{tabular}{@{}lcc@{}}
\toprule
\multicolumn{1}{c}{Method} & Accuracy             & Robustness           \\ \midrule
HEP                      & 37.44                & 22.35                \\
AVP                    & 37.62                & 25.04                \\
PAFT                    & 50.14          & \uline{39.14}       \\
APT-UC                     & \uline{58.38}       & 38.32          \\
APT-CSC                   & \textbf{60.21} &  \textbf{39.26} \\ \bottomrule
\end{tabular}
\end{table}

As shown in \cref{tab: rn50}, the results of our method with the ResNet50 vision encoder demonstrates a performance advantage consistent with that achieved based on ViT. Our method exhibits substantial enhancements over the zero-shot baseline. Furthermore, our approach attains robustness comparable to that achieved by LP, while achieving significantly higher accuracy.

\subsection{Ablation Study}
\label{app: ablation study}

The ablation study was conducted using the following set-up unless otherwise specified. 
The number of context vectors, $M$, was 16, the class embedding was placed at the \textit{end} of the prompt, the prompting strategy for training the adversary was \textit{on-the-fly}.
Models were trained with 16 shots for $\epsilon = 4/255$.

\subsubsection{Prompt Context Length}
\label{sec: prompt context length}
\begin{table}[]
\centering
\caption{The average performance of our methods with different number of context vectors, $M$. $\epsilon=4/255$ and the number of shots is 16.}
\label{tab: ablation context length}
\begin{tabular}{@{}ccccc@{}}
\toprule
{\multirow{2}{*}{$M$}} & \multicolumn{2}{c}{UC}                      & \multicolumn{2}{c}{CSC}                     \\ \cmidrule(l){2-5} 
  & Acc.                 & Rob.                 & Acc.                 & Rob.                 \\ \midrule
1       & 43.63                & 16.75                & 53.77                &  \textbf{20.77} \\
4       & 48.09                & 18.59                & 54.07                & 20.53                \\
8       & 50.19                & 19.65                & \textbf{54.21}                & 20.52                \\
16      &  \textbf{51.06} &  \textbf{20.19} &  \textbf{54.21} & 20.68                \\ \bottomrule
\end{tabular}
\end{table}

As shown in \cref{tab: ablation context length}, it is notable that as the number of context vectors increases, a significant enhancement in performance becomes evident for variant UC.
In the case of CSC, while there is a marginal increase in accuracy associated with the number of context vectors, there is no improvement in robustness. 
A longer context has a greater number of parameters and, consequently, it is expected to possess a larger capacity for improved performance. 
However, the observation that robustness is not notably improved for CSC may be attributed to a fact that even when the context length is one (the minimum), CSC inherently incorporates more parameters than UC with a context length of 16, provided that the number of classes exceeds 16. 
This phenomenon exists in the majority of our test datasets. The default context length was set to 16 based based on there results, due to its better performance.

\subsubsection{Class Embedding Position}
\label{sec: ablation study class embedding position}
\begin{table}[]
\centering
\caption{The average performance of our methods with the class embedding located at the different positions. $M=16$, $\epsilon=4/255$ and 16 shots.}
\label{tab: ablation class embedding position}
\begin{tabular}{@{}lcccc@{}}
\toprule
\multicolumn{1}{c}{Class Embedding} & \multicolumn{2}{c}{UC}                      & \multicolumn{2}{c}{CSC}                     \\ \cmidrule(l){2-5} 
\multicolumn{1}{c}{Position}   & Acc.                 & Rob.                 & Acc.                 & Rob.                 \\ \midrule
front                          & 51.16                & 20.04                & 54.09                & 20.56                \\
middle                         &  \textbf{51.91} & 20.16                &  \textbf{54.21} &  \textbf{20.68} \\
end                            & 51.06                &  \textbf{20.19} &  \textbf{54.21} & \textbf{20.68} \\ \bottomrule
\end{tabular}
\end{table}

The performance of our method was evaluated with the class embedded at different locations in the context (\cref{tab: ablation class embedding position}). This experiment reveals that no statistically significant differences are observed for both UC and CSC. As a result, \textit{end} was selected as the default position due to its slightly better performance and simplicity.

\subsubsection{Prompting for Training Adversarial Generation}
\label{sec: ablation adversarial generation strategy}
\begin{table}[]
\centering
\caption{The average performance of our methods using different prompting strategies for generating training adversarial examples. $M=16$, $\epsilon=4/255$ and the number of shots is 16.}
\label{tab: adv prompt}
\begin{tabular}{@{}lccccc@{}}
\toprule
\multicolumn{1}{c}{\multirow{2}{*}{$\bm{t}'$}} & \multirow{2}{*}{$\alpha'$} & \multicolumn{2}{c}{UC} & \multicolumn{2}{c}{CSC} \\ \cmidrule(l){3-6} 
\multicolumn{1}{c}{} &       & Acc.                & Rob.               & Acc.                & Rob.               \\ \midrule
\textit{constant}             & -     & 44.42                & 12.30                & 43.52                & 11.13                \\
\textit{on-the-fly}              & -     & 51.06                & {\ul20.19}       & 54.21                & {\ul20.68}       \\
\textit{perturbed}            & 0.1   &  \textbf{53.31} & 17.99                & \textbf{55.10} & 19.42                \\ \textit{perturbed}            & 0.01  & {\ul 52.01}       & 19.99                & {\ul54.25}       & 20.66                \\
\textit{perturbed}            & 0.001 & 51.60                &  \textbf{20.25} & 54.24                &  \textbf{20.69} \\
\bottomrule
\end{tabular}
\end{table}

\cref{tab: adv prompt} compares the performance of different variants of our method when training adversarial examples are generated by the different prompting strategies introduced in \cref{sec: method context optimization}.
We observe that the prompting strategies \textit{on-the-fly} and \textit{perturbed}, if $\alpha'$ is properly configured, achieve very close accuracy and robustness, while \textit{perturbed} seems to be slightly advantageous regarding accuracy under the unified context. 
Both of them perform much better than the strategy \textit{constant}.
Inside the strategy \textit{perturbed}, the perturbation magnitude increases as $\alpha'$ increases so that the perturbed prompts deviate more from the original ones, which results in the drop of performance as shown in the data. 

\subsection{Verification}
\label{sec: verification on adversarial evaluation}

\begin{table}[]
\centering
\caption{The average robustness under various attacks. The results of our methods are reported on the checkpoints trained on all shots. $M=16$ and $\epsilon=4/255$.}
\label{tab: more attacks eval}
\resizebox{\columnwidth}{!}{%
\begin{tabular}{@{}lccccc@{}}
\toprule
\multicolumn{1}{c}{Method} & Accuracy & PGD                  & TPGD  & CW    & AutoAttack \\ \midrule
HEP                  & 32.98 & 10.34                & 23.34 & 10.02 & 8.55       \\
APT-UC                    & 54.95 & 24.76                & 39.40 & 22.17 & 19.69      \\
APT-CSC & \textbf{59.43} & \textbf{27.05} & \textbf{42.80} & \textbf{24.69} & \textbf{22.47} \\ \bottomrule
\end{tabular}%
}
\end{table}

\begin{table}[]
\centering
\caption{The average robustness of our methods  against attacks based on the different attack prompts (columns). The \textbf{lowest robustness} is highlighted for each variant of our method (row). ``Learned" denotes the prompts learned by our method. P1-P6 are defined in \cref{fig: robustness varied prompts}. The results of our methods are reported for the checkpoints trained on all shots. $M=16$ and $\epsilon=4/255$.}
\label{tab: adv eval more prompts}
\resizebox{\columnwidth}{!}{%
\begin{tabular}{@{}lcccccccc@{}}
\toprule
\multicolumn{1}{c}{\multirow{2}{*}{Method}} & \multirow{2}{*}{Accuracy} & \multicolumn{7}{c}{Robustness} \\ \cmidrule(l){3-9} 
\multicolumn{1}{c}{} &       & Learned              & P1    & P2    & P3    & P4    & P5    & P6    \\ \midrule
APT-UC            & 54.95 & \textbf{24.76} & 35.78 & 35.30 & 35.37 & 35.26 & 37.92 & 36.58 \\
APT-CSC           & 59.43 & \textbf{27.05} & 42.43 & 42.09 & 42.10 & 42.04 & 44.17 & 43.35 \\ \bottomrule
\end{tabular}%
}
\end{table}

To verify that our evaluation of adversarial robustness is reliable, we first additionally evaluated the adversarial robustness of our methods using a diverse set of attacks including TPGD \citep{zhang_theoretically_2019}, CW \citep{carlini_towards_2017} and AutoAttack \citep{croce_reliable_2020}.
AutoAttack, particularly, is widely accepted as a reliable attack.
Note that for AutoAttack only we reduce the size of the ImageNet test set to 5000 following the practice used in RobustBench \citep{croce_robustbench_2021}, since running AutoAttack on the full ImageNet test set is very expensive. 
As shown in \cref{tab: more attacks eval}, the performance of our methods degrade by a reasonable amount but still improve over the baseline method by a large margin under the stronger attacks CW and AutoAttack. This suggests that the robustness advantage of our methods is not a consequence of overfitting to the particular attack or to obfuscated gradients \citep{athalye_obfuscated_2018}. 

To further exclude the possibility of our method masking the gradients for the particular attack prompts \citep{athalye_obfuscated_2018}, we evaluated our methods using adversarial examples transferred from other prompts with the same model as defined in \cref{fig: robustness varied prompts}.
It is shown in \cref{tab: adv eval more prompts} that our method achieves higher robustness against the adversarial examples generated by six other different prompts. 
This suggests that the robustness achieved by our method is not specific to the attack from a particular prompting source and can generalize well to attacks using different prompting methods.

\subsection{Interpretability of the Learned Context}
\label{app: interpreting learned contexts}
\begin{table}[]
\centering
\caption{The nearest word for each unified context vector learned by our method with 16 shots on different datasets. N/A means non-Latin characters.}
\label{tab: context vis}
\resizebox{\columnwidth}{!}{%
\begin{tabular}{@{}ccccccc@{}}
\toprule
\multirow{2}{*}{No.} & \multicolumn{2}{c}{ImageNet} & \multicolumn{2}{c}{Food101} & \multicolumn{2}{c}{DTD} \\ \cmidrule(l){2-7} 
   & Word         & Distance & Word       & Distance & Word         & Distance \\ \midrule
1  & optimizing   & 1.30     & yourselves & 1.18     & vid          & 0.82     \\
2  & dent         & 1.63     & aux        & 0.91     & N/A          & 0.96     \\
3  & daisies      & 1.07     & ester      & 0.94     & industries   & 0.67     \\
4  & hashtags     & 1.06     & blocker    & 0.82     & lala         & 0.98     \\
5  & ents         & 1.01     & obe        & 1.26     & tice         & 0.97     \\
6  & sph          & 1.14     & each       & 0.94     & met          & 1.02     \\
7  & swirl        & 1.04     & legally    & 1.24     & stan         & 1.27     \\
8  & ridge        & 1.20     & frock      & 1.21     & washingtondc & 1.33     \\
9  & respectively & 1.10     & notte      & 0.92     & press        & 1.31     \\
10 & stigma       & 1.02     & ur         & 0.91     & ely          & 1.52     \\
11 & spike        & 1.09     & offic      & 1.17     & Ĺ            & 1.18     \\
12 & pas          & 0.83     & councillor & 1.17     & soil         & 0.85     \\
13 & conquering   & 1.40     & dgers      & 1.17     & popefrancis  & 2.21     \\
14 & turk         & 1.19     & charging   & 1.35     & honourable   & 1.49     \\
15 & compares     & 1.21     & sk         & 1.09     & perth        & 1.50     \\
16 & artillery    & 1.39     & mindy      & 1.38     & methodology  & 1.76     \\ \bottomrule
\end{tabular}%
}
\end{table}

This section studies the interpretability of the context vectors learned by APT through the lens of the nearest words. 
The nearest word is, following the implementation of \citet{zhou_learning_2022}, the vocabulary whose embedding vector is closest to the learned vector based on the Euclidean distance. 
\cref{tab: context vis} shows the decoded nearest words for sixteen learned, unified, vectors ($M=16$) on the selected datasets. 
We find that, in contrast to the hand-engineered prompt contexts (\cref{tab: dataset}), the decoded nearest words appear to lack semantic relevance with the respective dataset.
It seems that APT may have learned something beyond current human vocabulary in order to defend against human-imperceptible  adversarial perturbations. 
On the other hand, the nearest words, despite being used in the previous works like \citep{zhou_learning_2022}, may not accurately reflect the semantic meaning of the learned vectors. 
To be more specific, there is a lack of well-established distance threshold above which the decoded words can be considered as faithful interpretation. 
Besides, some decoded words are non-Latin characters, \eg, the second word for DTD in \cref{tab: context vis}, which are uninterpretable to human.

\subsection{Dependency on the Robust Visual Backbone}
\label{app: dependency on robust backbone}

This section discusses the dependency of APT on the pre-trained adversarially robust image encoder. 
It was observed in our experiments that APT failed to improve robustness over the HEP baseline, or even collapsed during training, when the image encoder of CLIP is not pre-trained using a robust training methods like TeCoA \citep{mao_understanding_2023} to attain adversarial robustness. 
This suggests that a robust visual backbone is necessary for APT to be effective. 

We note that it was reported in \citet{chen_visual_2023} that AVP achieved 27.7\% for accuracy and 16.7\% for robustness against PGD100 with a standardly-trained ResNet18 \citep{he_deep_2016} on CIFAR10. 
The accuracy is dramatically compromised when compared to 94.9\% accuracy of the pre-trained model without AVP. 
Although the robustness increases from 0 to 16.7\%, the trade-off between accuracy and robustness seems unacceptable. 

\end{document}